# An LLM-based Two-Stage Transformer Framework for Cross-Domain Bearing Fault Diagnosis with Limited Data

Jinghan Wang *Student Member, IEEE*, Feng Cheng, Wentao Wu, Hang Li, Gaoliang Peng* *Member, IEEE* and Tianchen Liu *Member, IEEE*

***Abstract*: Bearing fault diagnosis faces critical challenges when dataset heterogeneity, operating condition variations, and limited labeled data occur simultaneously in industrial environments. Existing approaches address these issues in isolation and rely on implicit feature alignment, limiting effectiveness under concurrent challenges. This paper proposes a knowledge-guided two-stage transfer learning framework that employs a lightweight GPT-2-style Transformer with causal self-attention for hierarchical feature extraction from vibration signals, establishing explicit pathways where pre-trained encoder weights and fault prototype embeddings serve as knowledge carriers from multi-source pre-training to target adaptation. The framework addresses the dual-shift challenge through multi-source learning for generalizable representations, prototype-based knowledge modulation for target adaptation, and taxonomy-adaptive classification for seamless transfer across heterogeneous fault categories. Experimental validation on four real-world datasets demonstrates 92.61% average accuracy with only 10% labeled target data, outperforming state-of-the-art methods by 17.24 percentage points, establishing a practical pathway toward cost-effective predictive maintenance in Industry 4.0 applications.**

## I. Introduction

Bearings are critical components in rotating machinery across aerospace, manufacturing, and energy sectors, where failures account for approximately 40-50% of mechanical breakdowns and incur substantial downtime costs [1]. Accurate fault diagnosis is essential for predictive maintenance in Industry 4.0 applications [2]. Traditional signal processing methods rely on hand-crafted features and domain expertise, limiting their adaptability to diverse operating conditions [3]. While deep learning approaches have demonstrated high accuracy under controlled laboratory settings [4], [5], their deployment in industrial environments faces three concurrent challenges: i) equipment operates under time-varying conditions, ii) fault samples are scarce due to high reliability requirements, and iii) diagnostic models must generalize across different bearing types with heterogeneous sensor configurations. These challenges demand intelligent diagnosis methods capable of robust performance under simultaneous dataset shifts, operating condition variations, and limited labeled data.

Despite the promise of deep learning for addressing these limitations, existing methods face critical limitations in practical deployments. Deep learning methods including convolutional neural networks achieve high accuracy when training and testing data share similar distributions [6]. However, their performance degrades substantially under domain shifts caused by varying operating conditions, bearing geometries, or sensor specifications [7]. To address these limitations, transfer learning approaches such as domain adaptation [8], [9], [10] target cross-domain generalization, while meta-learning [11] and data augmentation [12] focus on small-sample scenarios. Despite these advances, existing methods usually address either cross-domain transfer or limited data learning in isolation, leaving a critical gap: the simultaneous dual-shift challenge where both dataset heterogeneity and operating condition variations occur with minimal target labels. Furthermore, most approaches lack explicit mechanisms to transfer knowledge across domains; instead, they rely on implicit feature alignment or requiring extensive labeled data from target domains [13]. This limitation hinders practical deployment where annotation costs are prohibitive and diverse bearing types must be diagnosed using knowledge learned from different source platforms.

Recent advances in transformer architectures have opened new possibilities for cross-domain transfer learning through their capabilities in knowledge retention and pattern recognition [14], [15]. The self-attention mechanism naturally captures temporal dependencies in sequential data, making transformers well-suited for processing vibration signals [16]. Recent transformer-based industrial diagnostics include self-supervised reconstruction with weakly supervised classification [17], multi-model collaboration for few-shot fault prediction [18], and feature textualization approaches [19]. While validating transformer applicability, existing approaches exhibit critical limitations in knowledge utilization and transfer mechanisms. Specifically, stages are trained independently without explicit knowledge transfer, failing to leverage universal fault patterns or simultaneously address cross-dataset and cross-condition challenges. Moreover, knowledge transfer relies on implicit feature alignment rather than explicit guidance, limiting effectiveness under severe data scarcity. These limitations reveal a fundamental challenge: how to establish explicit knowledge pathways enabling effective transfer when dataset heterogeneity, operating condition variations, and minimal target labels occur concurrently.

To address this gap, this paper proposes a knowledge-guided two-stage transfer learning framework establishing explicit pathways for transferring universal fault patterns from

*This work is supported by National Natural Science Foundation of China under Grant 52275099.

Jinghan Wang, Feng Cheng, Wentao Wu, Hang Li, Gaoliang Peng* and Tianchen Liu are with the School of Mechatronics Engineering, Harbin Institute of Technology, Harbin 150001, China (corresponding email: pgl7782@hit.edu.cn). *Corresponding author.

multi-source pre-training to target adaptation. The main contributions of this work are as follows.

- We propose an explicit knowledge-guided framework addressing the dual-shift scenario where dataset heterogeneity and operating condition variations occur simultaneously with limited labeled data.
- Explicit knowledge transfer mechanisms through parameter-level initialization and feature-level modulation are introduced, establishing concrete pathways that outperform implicit alignment under data scarcity.
- A dynamic classification head is developed that automatically adapts between heterogeneous fault taxonomies, enabling seamless cross-dataset transfer.

The rest of this paper is organized as follows: Section II presents the methodology, Section III reports experimental results, and Section IV concludes the paper with future directions.

## II. PROPOSED METHODOLOGY

This section presents our proposed knowledge-guided two-stage transfer learning framework, which is designed for cross-domain fault diagnosis scenarios under limited labeled data. Specifically, Section II-A formulates the dual-shift transfer problem. Section II-B overviews the framework architecture. Sections II-C and II-D detail the two stages.

### A. Problem Formulation

Consider a cross-domain fault diagnosis scenario with multiple source datasets $D_s = \{D_1, D_{2,}, \ldots D_m\}$ and a target dataset $D_t$. Each dataset $D_i$ comprises $N_i$ vibration signal-label pairs. For each sample $(x, y) \in D_i$, $x \in R^L$ represents a signal of length $L$ and $y \in Y_i$ denotes its fault category from the taxonomy $Y_i$ associated with dataset $D_i$.

Unlike conventional transfer learning addressing either domain shift or condition shift in isolation, industrial bearing diagnosis encounters simultaneous dual shifts. Dataset shift arises from heterogeneous bearing geometries, sensor configurations, and sampling rates, manifesting as divergent data distributions. Condition shift stems from varying operating speeds, loads, and environmental factors, leading to different signal-fault relationships. Additionally, taxonomy heterogeneity occurs when $|Y_s| \neq |Y_t|$, where $Y_s$ and $Y_t$ denote the fault category sets of source and target domains respectively, such as 3-class scheme (Normal, Inner Race Fault, Outer Race Fault) versus 4-class scheme that additionally includes Ball Fault.

Our objective is to learn a diagnostic function $f_\theta : R^L \rightarrow Y_t$ that achieves high accuracy on $D_t$ by transferring knowledge from $D_s$ while requiring only limited labeled target samples.

### B. Overall Framework

Our framework addresses the dual-shift challenge through a two-stage architecture with explicit knowledge transfer mechanisms, as illustrated in Fig. 1. The design philosophy is to decouple universal fault pattern learning from domain-specific adaptation, enabling efficient knowledge reuse under limited target labels.

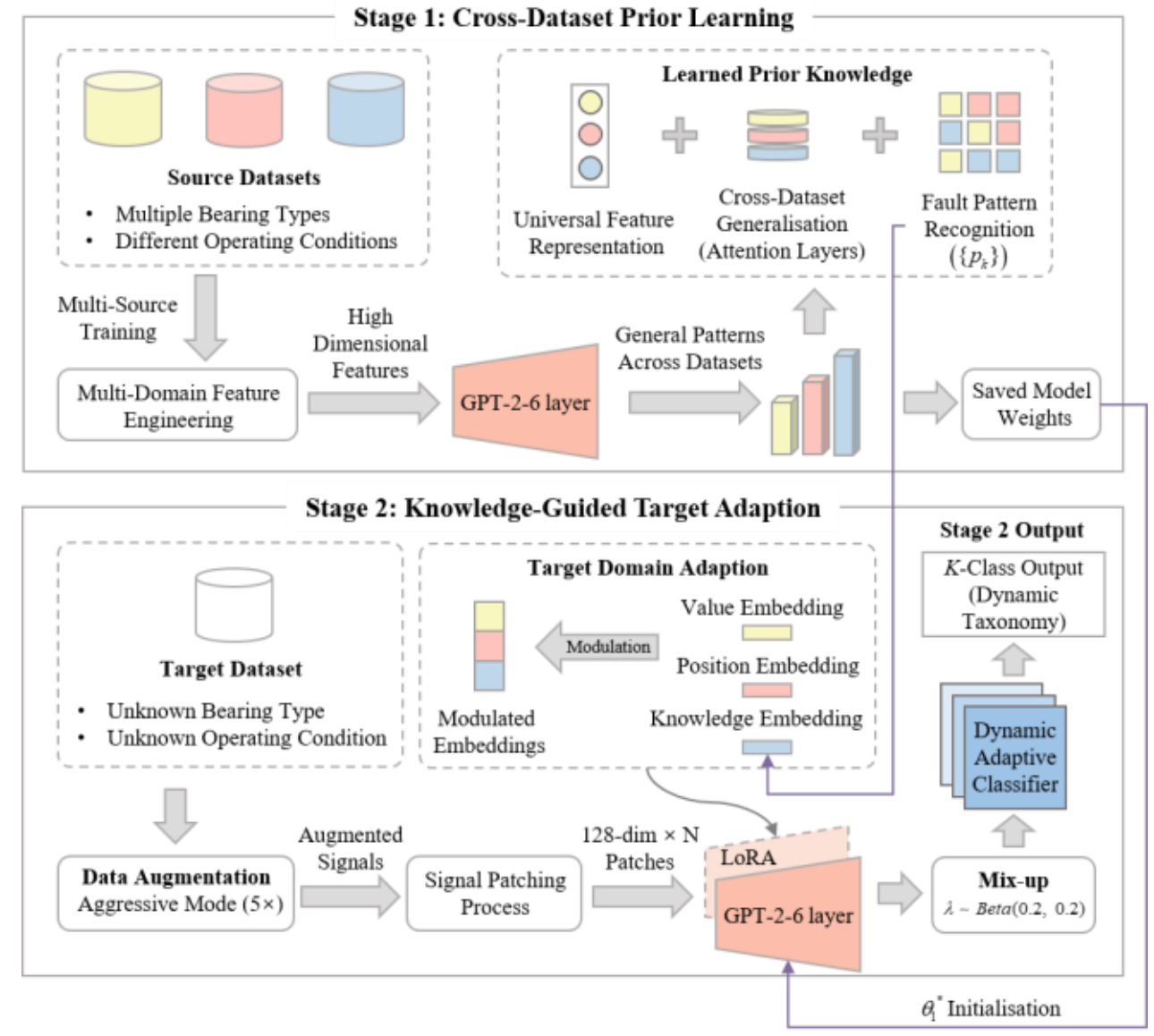

Figure 1. The overall framework of knowledge-guided two-stage transfer learning for cross-domain bearing fault diagnosis.

Stage 1, referred to as cross-dataset prior learning, learns generalizable fault representations from multiple source datasets $D_s$ using a lightweight GPT-2 encoder. The trained model produces two forms of transferable knowledge: encoder parameters $\theta_1^*$ encoding universal fault signatures, and $K$ fault prototype embeddings $\{p_k\}$ (for $k=1$ to $K$) representing characteristic patterns for each fault type extracted by averaging final token hidden states within each class.

Stage 2, referred to as knowledge-guided adaptation, adapts the learned prior to the target domain $D_t$ via dual-level knowledge transfer. At the parameter level, the encoder in Stage 2 initializes with $\theta_1^*$, providing a strong starting point. At the feature level, fault prototypes $\{p_k\}$ are injected into the input embedding layer, guiding the model to focus on fault-relevant signal regions through prototype-similarity weighting. Unlike methods that train stages independently, our framework establishes explicit knowledge pathways through both weight inheritance and feature-space guidance, enabling effective adaptation with minimal target labels. Additionally, Stage 2 incorporates parameter-efficient fine-tuning via LoRA, a dynamic classification head for heterogeneous taxonomies, and strategic data augmentation.

### C. Stage 1: Cross-Dataset Prior Learning

Stage 1 extracts fault representations that generalize across diverse bearing types and operating conditions through multi-source feature extraction and transformer-based encoding.

Raw vibration signals from all source datasets are transformed into high dimensional feature vectors capturing multi-scale fault signatures. This comprehensive feature set ensures robustness to varying sampling rates and heterogeneous sensor configurations. The high dimensional features are encoded using a 6-layer GPT-2 model with hidden dimension 768, employing 12 attention heads and feedforward dimension 3072, totaling approximately 43 million parameters. Input features undergo linear projection from 81 dimensions to 768 dimensions. The model outputs 768-dimensional representations extracted from the final token's hidden state, which serves as the sequence-level fault representation. We adopt GPT-2 decoder-only architecture for its strong feature extraction capabilities transferable from linguistic pretraining and unidirectional attention reducing inference complexity critical for industrial deployment. The 6-layer depth balances representational capacity with parameter efficiency. The term "LLM-based" in this paper refers to the structural and training paradigm inherited from GPT-2, including its causal self-attention mechanism, positional encoding, and autoregressive pre-training objective, rather than implying billion-parameter scale or natural language modality. Specifically, the implemented model is a lightweight 6-layer GPT-2-style Transformer with approximately 43 million parameters, optimized for vibration signal processing in industrial deployment. Stage 1 is trained via supervised learning over all source datasets using cross-entropy loss:

$$L_{S1} = \left(\frac{1}{|D_s|}\right) \sum\nolimits_{\{(x,y)\in D_s\}} L_{CE}\left(f_{\theta_1}(x), y\right), \tag{1}$$

where $L_{CE}$ denotes the cross-entropy loss function, and $f_{\theta_1}$ denotes the Stage 1 model with parameters $\theta_1$. Multi-source training learns dataset-invariant representations addressing the dataset shift component.

Upon convergence, Stage 1 produces two forms of transferable knowledge. First, encoder weights $\theta_1^*$ encode universal fault patterns in attention and feedforward layers, initializing Stage 2 encoder. Second, fault prototype embeddings $\{p_k\}$ for *k=1* to *K* are extracted by averaging final token hidden states within each class:

$$p_k = \left(\frac{1}{|S_k|}\right) \sum\nolimits_{\{x\in S_k\}} h_{\theta_1}(x), \tag{2}$$

where $S_k = \{x : y = k,\ (x, y) \in D_s\}$ denotes samples of fault type *k*, and $h_{\theta_1}(x)$ represents the final token's hidden state from the converged model. These prototypes encapsulate characteristic patterns for each fault category and guide attention mechanisms in Stage 2.

### D. Stage 2: Knowledge-Guided Adaptation

Stage 2 adapts universal fault knowledge to the target domain using only 10% labeled data through four synergistic mechanisms: data augmentation, prototype-modulated embeddings, parameter-efficient fine-tuning, and dynamic classification.

Given the scarcity of labeled target samples, we employ aggressive augmentation strategies. Signal-level augmentation provides a fivefold increase in training samples via three operations: Gaussian noise injection, time scaling, and circular time shifts. For noise injection, the augmented signals are generated as:

$$\tilde{x} = x + \varepsilon, \tag{3}$$

where $\varepsilon$ is drawn from $N(0,\ \sigma^2)$ with $\sigma = 0.01 \cdot std(x)$, ensuring noise amplitude adapts to individual signal magnitudes. Time scaling applies random factors uniformly sampled from [0.9, 1.1], and circular time shifts introduce random offsets up to 10% of signal length.

Mix-up regularization further enhances generalization by generating virtual training samples through linear interpolation of randomly selected sample pairs:

$$\tilde{x} = \lambda x_i + (1-\lambda) x_j,\ \tilde{y} = \lambda y_i + (1-\lambda) y_j, \tag{4}$$

where $(x_i,\ y_i)$ and $(x_j,\ y_j)$ are two randomly drawn training samples, and $\lambda \sim Beta(0.2,\ 0.2)$. Signal-level augmentation expands the training set by $5\times$, while Mix-up regularization continuously generates diverse sample combinations during training, jointly mitigating overfitting under limited labeled data.

Unlike Stage 1 operating on high-dimensional features, Stage 2 processes raw signals to capture fine-grained temporal patterns. However, raw signals lack semantic structure, which we address by modulating input embeddings with fault prototypes in Stage 1. Each raw signal *x* of length *L* is partitioned into *N* = *L/128* non-overlapping patches $\{s_1, s_2, \ldots, s_N\}$ where each patch $s_N$ has dimension 128.

Patch embeddings comprise three components. Value embedding projects each patch into 768-dimensional space:

$$e_i^{val} = W_v s_i + b_v, \tag{5}$$

where $W_v$ is a 768×128 matrix and $b_v$ is a 768-dimensional bias vector. Position embedding preserves temporal ordering:

$$e_i^{pos} = PE(i), \tag{6}$$

where *PE(i)* is a trainable 768-dimensional position-dependent vector. Prototype modulation, the core innovation of Stage 2, computes similarity scores between each patch embedding and all *K* fault prototypes using scaled dot-product attention:

$$\alpha_{\{i,k\}} = \frac{\exp(e_i^{val} \cdot \frac{p_k}{\sqrt{d}})}{\sum_{k'=1}^{K} \exp(e_i^{val} \cdot \frac{p_{\{k'\}}}{\sqrt{d}})}, \tag{7}$$

where *d* equals 768. These scores indicate which fault types are most relevant to the current patch. We construct a prototype-weighted modulation vector:

$$m_i = \sum\nolimits_{k=1}^{K} \alpha_{\{i,k\}} p_k, \tag{8}$$

The final embedding combines all components via element-wise multiplication and addition:

$$e_i = LayerNorm(m_i \odot (e_i^{val} + e_i^{pos})), \tag{9}$$

where $\odot$ denotes element-wise multiplication. This prototype modulation acts as a learnable attention mask amplifying embeddings of patches resembling known fault patterns from Stage 1, accelerating convergence, and improving data efficiency when only a few target labels are available.

The modulated embeddings are fed into a 6-layer GPT-2 encoder initialized with Stage 1's weights $\theta_1^*$. To preserve this prior while adapting to target-specific patterns, we employ LoRA. For each attention weight matrix *W*, we inject trainable low-rank decomposition:

$$h = W_0 x + \Delta W x = W_0 x + \left(\frac{\alpha}{r}\right) BAx, \tag{10}$$

where $W_0$ is initialized from $\theta_1^*$ and remains frozen, A is an $r \times d$ matrix, B is a $d \times r$ matrix forming a rank-r decomposition. This approach achieves parameter efficiency where only approximately 0.5% of parameters are trainable, prior preservation where frozen $W_0$ retains universal representations, and fast convergence with limited data.

To accommodate heterogeneous fault taxonomies, we employ a taxonomy-adaptive classification head:

$$\hat{y} = \text{softmax}(W_c h_{final}), \tag{11}$$

where $W_c$ is a $K \times 768$ matrix with output dimension $K$ determined by the target taxonomy, and $h_{final}$ represents the hidden state of the last token from the GPT-2 encoder. The classification head is randomly initialized since source and target taxonomies may differ.

## III. Experimental Investigations

This section validates the proposed framework through comprehensive experiments on four real-world bearing datasets. Section III-A describes the experimental protocol and datasets. Section III-B evaluates knowledge-guided transfer under varying data availability. Section III-C analyzes computational efficiency.

### A. Experimental Protocol

Four publicly available bearing fault diagnosis datasets are employed for evaluation, as detailed in Table I. The datasets exhibit substantial heterogeneity in bearing types, sampling rates ranging from 12 kHz to 97.6 kHz, operating conditions, and fault taxonomies encompassing 3-class and 4-class scenarios. The heterogeneous sensor specifications spanning this wide sampling rate range necessitate robust feature extraction in Stage 1 to ensure knowledge transferability across domains.

TABLE I. Dataset Characteristics

| Dataset | Bearing Type | Conditions | Classes | Samples | Sampling Rate |
|---|---|---|---|---|---|
| CWRU | SKF 6205 | 4 speeds | 4 | 7,653 | 12 kHz |
| MFPT | Custom | 1 speed | 3 | 4,860 | 97.6 kHz |
| JNU | NSK NU2205 | 3 loads | 4 | 7,901 | 50 kHz |
| PU | INA HFL1026 | 4 speeds | 3 | 33,724 | 25.6 kHz |

A 4-fold cross-validation protocol is employed where each dataset serves as target domain once while the remaining three constitute source domains, ensuring comprehensive evaluation across diverse domain shifts. The two-phase training pipeline proceeds as follows: Phase 1 pre-trains on three source domains, then Phase 2 adapts to the target domain. Two data allocation scenarios address distinct objectives. Scenario 1 allocates 70% target samples for Phase 2 adaptation and 30% for testing, validating the knowledge-guided mechanism through Stage 1 versus Stage 2 comparison. Scenario 2 allocates 10% for adaptation and 90% for testing, evaluating data efficiency in the critical industrial setting with minimal labeling. Within Scenario 2, both cross-domain transfer with source pre-training and single-domain learning without source pre-training are examined to isolate the contribution of cross-domain knowledge.

The proposed method is compared against [19], a state-of-the-art multi-source transfer learning method employing an LLM-based architecture. In the subsequent text, MSSP is used as the corresponding abbreviation. Both methods use identical datasets and transformer architectures, isolating the effect of explicit knowledge guidance.

Implementation uses PyTorch 2.0.1 on an NVIDIA RTX 4090 GPU with 24 GB memory. Stage 1 employs AdamW optimizer with learning rate $5 \times 10^{-5}$, batch size 32, and 10 epochs on three source domains. Stage 2 uses learning rate $2 \times 10^{-5}$, batch size 32. LoRA uses rank $r$=8 and scaling factor $\alpha$=16. Identical hyperparameters are used across all datasets.

### B. Knowledge-Guided Transfer Evaluation

#### 1) Knowledge-Guided Mechanism Validation

The first experiment validates whether the two-stage knowledge-guided architecture provides genuine performance gains beyond Stage 1's universal feature extraction.

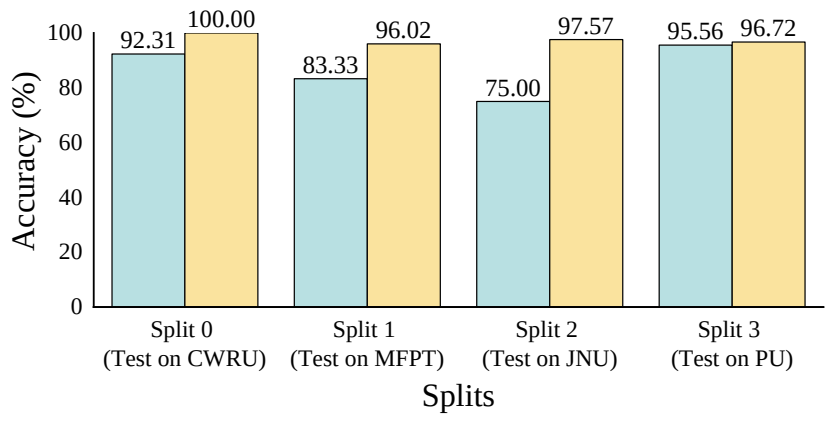

Figure 2. Performance comparison between Stage 1 and Stage 2. The green bars represent the Stage 1 performance while the yellow ones denote the Stage 2's.

TABLE II. Stage 2 Detailed Performance Metrics Under Scenario 1

| Target Dataset | Accuracy | Precision | Recall | F1-Score |
|---|---|---|---|---|
| CWRU | 100% | 1.0000 | 1.0000 | 1.0000 |
| MFPT | 94.70% | 0.8929 | 0.9470 | 0.9104 |
| JNU | 96.56% | 0.9670 | 0.9656 | 0.9655 |
| PU | 95.62% | 0.9577 | 0.9562 | 0.9569 |
| **Average** | **96.72%** | **0.9544** | **0.9672** | **0.9582** |

Stage 2 consistently outperforms Stage 1 across all four target datasets, achieving an average accuracy of 96.72% compared to Stage 1's 85.69%, representing an 11.03 percentage point improvement. As shown in Fig. 2 and Table

II, the knowledge-guided adaptation mechanism demonstrates substantial gains ranging from 1.16 % on MFPT to 22.57% on JNU. The latter result is particularly striking, indicating that explicit knowledge modulation through prototype-weighted embeddings and parameter-level initialization becomes increasingly valuable when the target domain exhibits complex load variations and significant distribution shift from source domains. The dynamic classification head seamlessly adapts to heterogeneous taxonomies, maintaining high performance across both 3-class and 4-class scenarios without architectural modifications. These results validate that the explicit knowledge pathways from Stage 1 to Stage 2 through parameter initialization and feature modulation provide genuine transfer learning benefits beyond universal feature extraction alone.

*2) Data Efficiency Evaluation: Cross-Domain Transfer*

The second experiment evaluates the critical industrial scenario under Scenario 2 where only 10% of target domain samples are available for Phase 2 adaptation.

The proposed framework achieves 92.61% average accuracy with only 10% labeled target data, outperforming MSSP by 17.24 percentage points, as shown in Fig. 3. This substantial improvement validates that explicit knowledge pathways through $\theta_1^*$ parameter initialization and $\{p_k\}$ prototype modulation enable effective learning under severe data scarcity. The framework demonstrates exceptional gains on datasets with complex fault manifestations. CWRU still attains perfect transfer at 100% accuracy. Notably, PU exhibits comparable performance between both methods around 85–87%, suggesting that PU's large-scale industrial bearing characteristics and substantial domain gap from laboratory source datasets create a challenging transfer scenario where even explicit knowledge guidance faces limitations. This performance gap is attributable to four compounding factors, including heterogeneous sampling rates across source and target domains (12–97.6 kHz vs. 25.6 kHz), inconsistent fault taxonomy schemes between laboratory and industrial settings, higher intra-class variability under variable-speed conditions, and substantial geometric differences between PU bearings (INA HFL1026) and source-domain bearings, which collectively amplify both the dataset shift and condition shift components of the dual-shift challenge. Future work should therefore investigate domain-specific pre-training or adaptive sampling strategies to bridge such multi-dimensional distributional gaps for industrial-grade bearing datasets. Fig. 4 reveals strong fault classification, with CWRU achieving perfect per-class separation and other datasets maintaining high diagonal concentration.

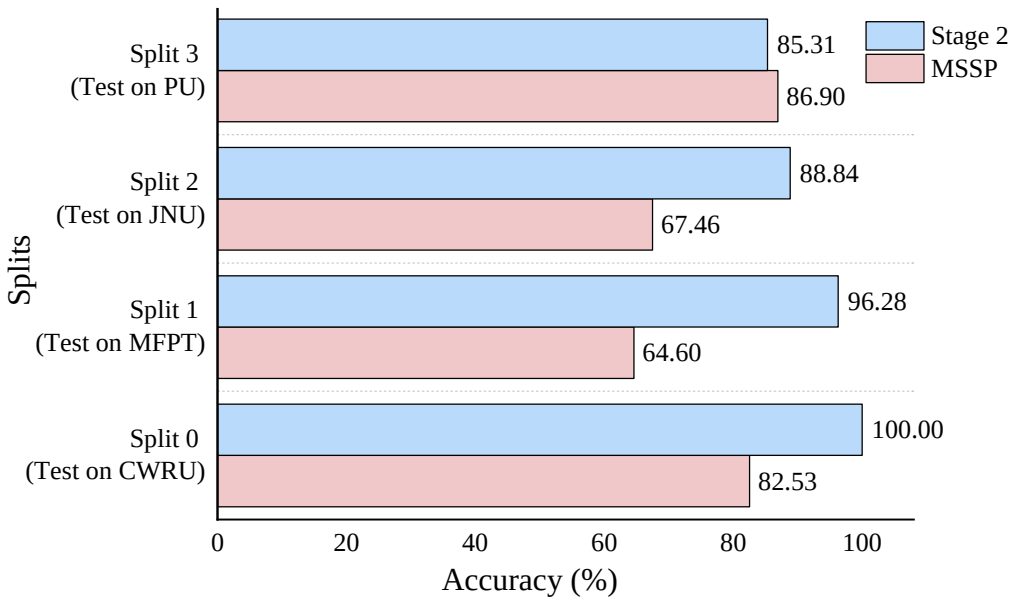

Figure 3. Cross-Domain Transfer Performance Comparison

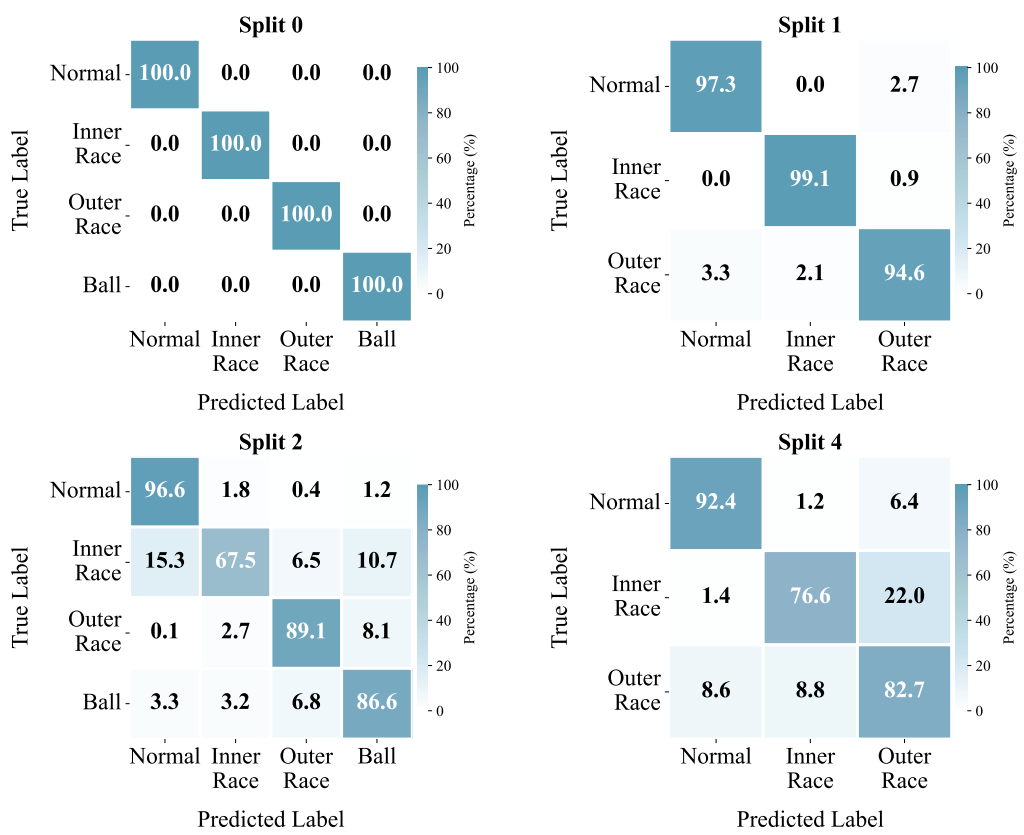

Figure 4. Confusion Matrices for Cross-Domain Transfer

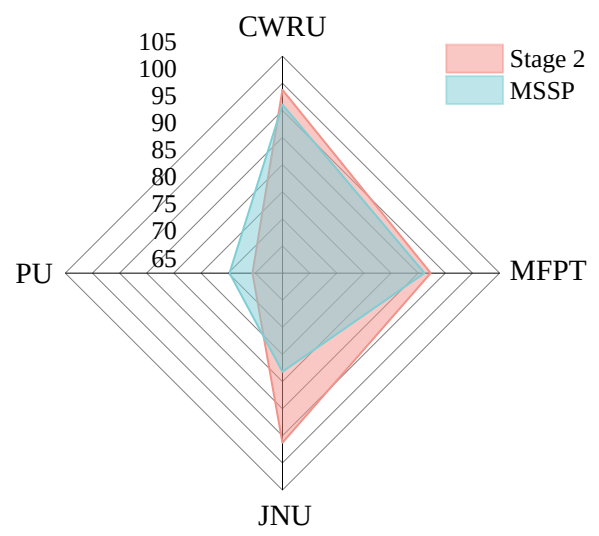

Figure 5. Single-Domain Limited Data Performance

*3) Data Efficiency Evaluation: Single-Domain Learning*

To isolate the contribution of cross-domain knowledge transfer, an ablation experiment is conducted under Scenario 2 where Phase 2 is trained directly on 10% target data without Phase 1 pre-training on source domains. This configuration evaluates whether the data efficiency benefits arise from explicit knowledge guidance mechanisms or merely from the two-stage architecture itself.

Radar chart comparing proposed method versus MSSP across four datasets under single-domain configuration. As shown in Fig. 5, the proposed method achieves 89.47% average accuracy even without source domain pre-training, outperforming MSSP by 3.18 percentage points. This validates that the architecture with prototype modulation and LoRA fine-tuning provides inherent data efficiency benefits beyond cross-domain knowledge transfer. Nevertheless, comparing with cross-domain results in Section III. B 2), leveraging source pre-training consistently delivers superior or comparable performance, with average accuracy improving from 89.47% to 91.18%. These findings confirm that while the proposed mechanisms enable effective learning with limited data, multi-source pre-training remains the recommended strategy for practical industrial deployment.

## C. Computational Efficiency Analysis

TABLE III. TRAINING TIME BREAKDOWN FOR TWO-STAGE FRAMEWORK

| Target Dataset | Stage 1 Source Training | Stage 1 Target Finetune | Stage 2 Source Training | Stage 2 Target Finetune |
|---|---|---|---|---|
| CWRU | 0.50 s/epoch | 0.16 s/epoch | 33 s/epoch | 3 s/epoch |

| MFPT | 0.18 s/epoch | 0.05 s/epoch | 37 s/epoch | 6 s/epoch |
|---|---|---|---|---|
| JNU | 0.18 s/epoch | 0.04 s/epoch | 39 s/epoch | 11 s/epoch |
| PU | 0.14 s/epoch | 0.11 s/epoch | 17 s/epoch | 49 s/epoch |

Table III presents the computational profile across four cross-dataset configurations. Stage 1 training on three source domains exhibits fast training ranging from 0.14 to 0.50 seconds per epoch, attributable to the compact 81-dimensional feature representation. Multi-source pre-training completes in approximately 5 minutes for 10 epochs. Stage 2 requires substantially more time ranging from 17 to 39 seconds per epoch due to raw signal patch processing, yet LoRA's parameter-efficient fine-tuning enables rapid target adaptation in 3 to 49 seconds per epoch depending on target dataset size. For the most demanding configuration on PU under Scenario 2 with 100 epochs, complete two-stage training within 1.5 hours. Critically, the two-stage architecture amortizes Stage 1 computational cost across multiple target domains, where a single Stage 1 pre-training on source bearings enables adaptation to numerous industrial sites with varying target bearing types and operating conditions through Stage 2 fine-tuning.

## IV. CONCLUSIONS

This paper presents an LLM-based knowledge-guided two-stage Transformer framework for cross-domain bearing fault diagnosis under simultaneous dataset shifts and operating condition variations with limited labeled data. Experimental results on four real-world datasets demonstrate that explicit knowledge pathways through parameter-level initialization and feature-level prototype modulation enable effective transfer learning with minimal supervision. The framework achieves 96.72% average accuracy with 70% target data and 92.61% with only 10% labeled samples, outperforming state-of-the-art methods by 17.24 percentage points in the limited data setting. The framework reduces annotation requirements by 90% while maintaining high diagnostic accuracy, with complete training requiring approximately 1.4 hours. These results establish that transformer-based architectures with explicit knowledge guidance can effectively address the dual-shift challenge in industrial fault diagnosis, providing a practical pathway toward cost-effective predictive maintenance across diverse rotating machinery.

While the proposed framework demonstrates substantial performance gains, it relies heavily on high-quality source data. Future work should investigate adaptive knowledge filtering to handle data quality issues, and extend the methodology to diverse sensing modalities beyond vibration signals.